\documentclass[conference]{IEEEtran}
\IEEEoverridecommandlockouts
\usepackage{cite}
\usepackage{amsmath,amssymb,amsfonts}
\usepackage{algorithmic}
\usepackage{graphicx}
\usepackage{textcomp}
\usepackage{xcolor}

\usepackage{bm} 

\def\BibTeX{{\rm B\kern-.05em{\sc i\kern-.025em b}\kern-.08em
    T\kern-.1667em\lower.7ex\hbox{E}\kern-.125emX}}
\begin{document}

\title{Introducing Echo Networks for Computational Neuroevolution\\
\thanks{The study was funded by the Bavarian Ministry of Economic Affairs, 
Regional Development and Energy within the Digital Signal Processing using 
Artificial Intelligence (DSAI) project.} 
}

\author{\IEEEauthorblockN{Christian Kroos*}
\IEEEauthorblockA{\textit{Audio \& Media Technologies} \\
\textit{Fraunhofer Institute for Integrated Circuits, IIS}\\
Erlangen, Germany \\
christian.kroos@iis.fraunhofer.de}
*Corresponding author
~\\
\and
\IEEEauthorblockN{Fabian K\"uch}
\IEEEauthorblockA{\textit{Audio \& Media Technologies} \\
\textit{Fraunhofer Institute for Integrated Circuits, IIS}\\
Erlangen, Germany \\
fabian.kuech@iis.fraunhofer.de}
}

% \author{\IEEEauthorblockN{Anonymous}
% ~\\
% \and
% \IEEEauthorblockN{Anonymous}
% }

\maketitle

\begin{abstract}
For applications on the extreme edge, minimal networks of only a few dozen artificial 
  neurons for event detection and classification in discrete time signals would be highly 
  desirable. Feed-forward networks, RNNs, and CNNs evolved through evolutionary algorithms 
  can all be successful in this respect but pose the problem of allowing little systematicity 
  in mutation and recombination if the standard direct genetic encoding of the weights 
  is used (as for instance in the classic NEAT algorithm). We therefore introduce 
  Echo Networks, a type of recurrent network that consists of the connection matrix 
  only, with the source neurons of the synapses represented as rows, destination neurons 
  as columns and weights as entries. There are no layers, and connections between 
  neurons can be bidirectional but are technically all recurrent. Input and output can be 
  arbitrarily assigned to any of the neurons and only use an additional (optional) function 
  in their computational path, e.g., a sigmoid to obtain a binary classification output. 
  We evaluated Echo Networks successfully on the classification of electrocardiography 
  signals but see the most promising potential in their genome representation as a single 
  matrix, allowing matrix computations and factorisations as mutation and 
  recombination operators.   
\end{abstract}


\section{Background}
\label{sec:background}
Devices on the extreme edge are characterized by tight energy use limitations
which restrict potential applications to minimal computational resources. For event detection 
and classification in discrete time signals on the extreme edge even small neural networks with a 
few thousand parameters tend to exceed the computation and/or memory limits. Accordingly, minimal networks
consisting only of a few dozen artificial neurons and performing just well enough on a given task
would be highly desirable. In most cases, however, no method exists to determine what kind of architecture 
and number of neurons and synapses are minimally needed. 

\begin{figure}[htbp]
\centerline{\includegraphics[trim=200 220 150 220,clip,scale=0.5]{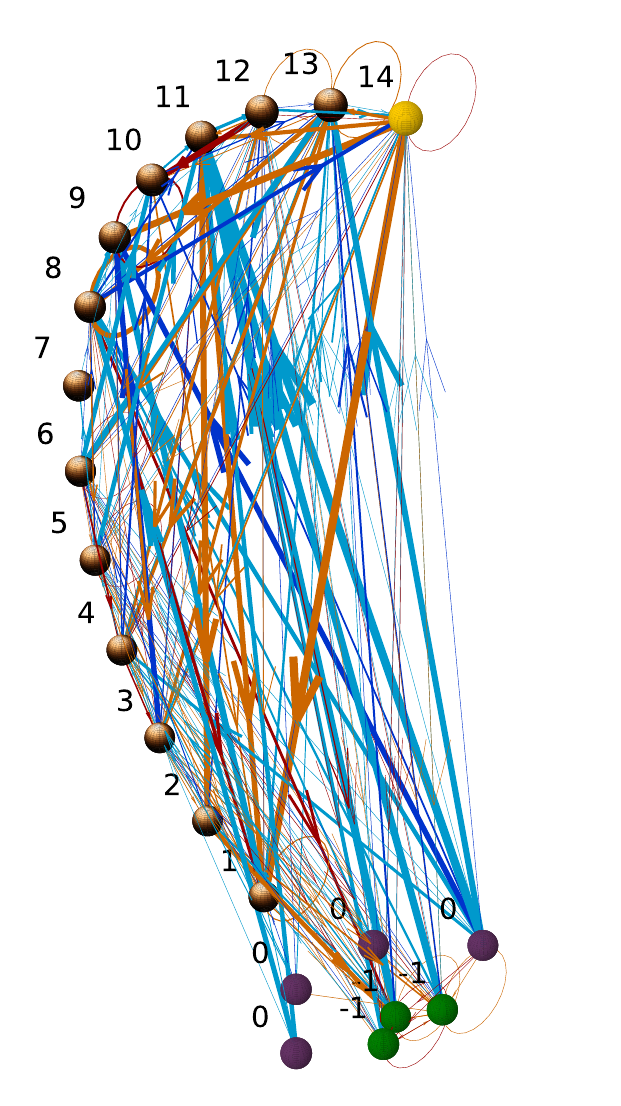}}
\caption{Sample of an RNN evolved through neuroevolution for the classification of
electrocardiography signals. The network reached an accuracy of $0.684$ on 
the test set in the ECG classification task (see section~\ref{sec:datasets}) consisting 
of $21$ neurons and $250$ weights.
Input neurons: green; output neuron: yellow; bias neurons: 
purple; hidden neurons: brown. Forward synapses: blue; recurrent synapses: red. Lighter
hues indicate a weight with a negative sign, darker hues a weight with a positive
sign. Line width is proportional to weight magnitude. Numbers close to the neurons indicate
assigned layers.}
\label{fig:sample_RNN}
\end{figure}

Neuroevolution \cite{stanley2019designing}, the population-based creation of artificial neural networks using Evolutionary Algorithms (EA)
and specifically Genetic Algorithms (GA), allows for the generation of tiny and highly task-adapted networks. In particular, 
the well-known NEAT algorithm (Neuroevolution of Augmenting Topologies, \cite{stanley2002evolving}) starts with 
the smallest possible network (connecting only input to output with no hidden neurons) and then grows it by 
adding neurons and synapses so long as the enlargement results in a performance improvement. NEAT goes beyond 
automated Neural Architecture Search (NAS) in also determining the weights of all synapses through random 
mutations and recombinations, thus, not requiring the computation of the gradient and backpropagation. NEAT uses 
direct encoding, i.e., the genetic representation specifies the network itself and does not 
consist of an abstract representation from which the network is generated. The independence from gradient 
estimation and backpropagation comes at the cost of a less efficient training process, although neuroevolution 
has been shown to be equivalent to gradient descend in the presence of Gaussian noise 
\cite{whitelam2021correspondence}. There is anecdotal evidence and theoretical 
research showing that minimal (usually under-parameterised) networks are difficult to train with backpropagation, presumably 
because the multi-dimensional loss landscape is less smooth exhibiting many problematic (high-value) isolated
local minima \cite{pmlr-v38-choromanska15}.

Regarding network type, evolved feed-forward networks and RNNs have been successfully
applied in many tasks. CNNs evolved through evolutionary 
algorithms also exist but usually only the architecture is determined via mutational variation, 
not the convolution coefficients themselves. All of these network types, however, pose the 
problem of allowing little systematicity in mutation and recombination of the weights: In the case of 
mutation, new weights are determined for a randomly selected subset of synapses, either by drawing 
randomly from a specified distribution or by perturbing the current weights, e.g., through Gaussian 
perturbation. In the case of recombination, for each synapse the weight is either taken from one of 
the parents (crossover) or the new weight is computed as the average of the parents' weights.

The procedure works well for small networks and requires only moderately sized populations 
of a few hundred individuals or even less. However, by specifying each weight separately 
evolving larger networks becomes problematic due to the computational effort involved given
the exponentially growing search space. Consequently, attempts have been made to create
larger patterns from limited genetic code, e.g. the compositional pattern producing 
networks of \cite{stanley2007compositional}.

The current study focuses on tiny networks, well-suited to the 
standard mutation and recombination operators. Here it is not network size but the lack of systematicity 
in the variation of the generated networks that is paramount. The generated networks appear 
to be almost always unique: different runs produce very different networks albeit often with comparable 
performance. In contrast to large networks with fixed structure and trained via backpropagation, these 
differences are substantial. While, for instance, in a billion parameter Large 
Language Model (LLM) it can be assumed that the weights of individual layers are only constrained to come from a specific 
but unknown distribution, the values of individual weights and the topology of the often
sparsely connected minimal networks are inherently meaningful. Each network resulting from
the augmenting evolution process represents a unique approach and often even a unique underlying 
solution (attempt). See Fig.~\ref{fig:sample_RNN} for a sample network evolved for the 
classification of electrocardiography signals.

Although the specificity of the resulting networks can be considered a strength in applications,
it is less advantageous when it comes to understand the solutions implemented by the network and, 
more importantly, if the evolution process fails repeatedly to generate a network within acceptable 
performance limits. Hyperparameter variations and additional constraints inserted into the fitness
formulation often help, but a more principled approach for mutation and recombination operators, 
while keeping direct encoding, would be preferable. We therefore introduce a new type of network 
named \textit{Echo Network}, which allows employing methods from matrix algebra to mutation and 
recombination.

\section{Proposed Concept}
\label{sec:concept_and_method}
\begin{figure}[htbp]
\centerline{\includegraphics{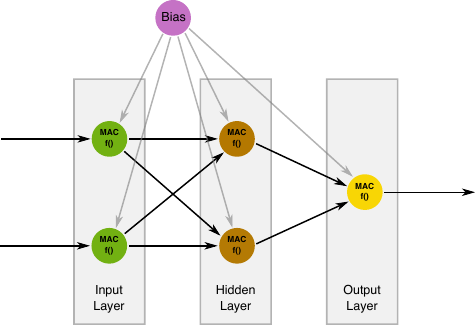}}
\caption{Schematic of a simple MLP.}
\label{fig:schematic_MLP}
\end{figure}

Echo Networks are recurrent neural networks where the topology and the weights are
fully represented by their connection matrix. The rows of the matrix define source neurons of 
the synapses of the network and the columns define destination neurons. The 
matrix entries are comprised of the weights with an exact zero indicating a missing connection
between two neurons. Using the connection matrix as a \textit{description tool} for networks is not 
novel (e.g., \cite{dejong2006evolutionary}). 
In an Echo Network, however, the connection matrix fully defines the network. It acts on the 
post-activation states of the previous \textit{evaluation step} of the network and not the 
input from the previous layer. Input and output can be arbitrarily assigned to any of the neurons. 
More formally, let us define a layer of a conventional MLP in the standard way as
\begin{equation}
  \bm{y}_{l} = f(\bm{W}_{l}^T \bm{x}_{l-1} + \bm{b}_{l}) 
\end{equation}
where $f$ is the activation function, $\bm{W}_{l}$ the weight matrix of the current layer, 
$\bm{b}_{l}$ the bias vector, and $\bm{x}_{l-1}$ the input from the previous layer or the general
input. Fig.~\ref{fig:schematic_MLP} shows a schematic of the well-known architecture.

\begin{figure}[htbp]
\centerline{\includegraphics{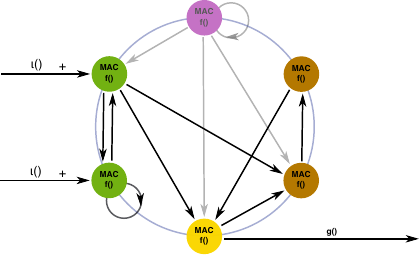}}
\caption{Schematic of a simple Echo Network}
\label{fig:schematic_echo_network}
\end{figure}

Accordingly, an Echo Network would then be defined by
\begin{equation}
  \bm{a}_{t} = f(\bm{C}^T \bm{a}_{t-1}) 
\end{equation}
with the bias term missing, $\bm{C}$ denoting the connection matrix and $\bm{a}_{t-1}$ 
denoting the post-activation state with the layer index being replaced by the (time) step index $t$, 
thus, $\bm{a}_{t-1}$ consists of the post-activations values resulting from the previous evaluation 
step of the network. At first glance, this looks like a very minor change, but it implies substantially 
different processing since the system is closed now. For the network to be useful, input from the 
outside has to be included by designating a subset of neurons $\mathcal{I}$ as input neurons and adding 
the input to the aggregation state before activation:  
\begin{equation}
  \bm{a}_{i,t} = f(\bm{c}_i^T \bm{a}_{i,t-1} + \iota_i( \bm{x}_{i,t}))
\end{equation}
where $\iota$ is an optional input function which might differ for different inputs, and
$i \in \mathcal{I}$.
Similarly, the output need to be extracted from the network from a set of designated output 
neurons $\mathcal{O}$:
\begin{equation}
  \bm{y}_{o,t} = g(\bm{c}_o^T \bm{a}_{o,t-1}) 
\end{equation}
where $g$ is a dedicated output function, e.g., in a binary classification task the sigmoid
function, and $o \in \mathcal{O}$. Note that $g$ has to be applied to the aggregation state before 
the applications of the neuron's own activation function. This guarantees that neurons can
be used arbitrarily as output neurons since their activation function does not need to produce 
values suitable for the final classification decision or regression task. 
Fig.~\ref{fig:schematic_echo_network} shows a schematic of an Echo Network.

There are no layers in Echo Networks, all neurons reside on the same level. Even the depiction 
of the neurons as lying on a circle does not accurately reflect their relationship since the 
'distance' between each pair of neurons is equal. 
There are, however, different processing path lengths due to the different numbers of 
processing steps until the output neuron (or any other neuron) is reached. 

The connection matrix enables bidirectional synapses between
neurons (upper and lower triangular partial matrices) and self-recurrent synapses of a neuron
(diagonal of the connection matrix). The distinction between forward and recurrent synapses
becomes less straightforward, though. On the one hand, technically, all synapses in an Echo 
Network are recurrent since the neurons always act on previous states of the same 
network (except for the first evaluation step after initialisation). On the other hand,
any computational path through the network equivalent to an acyclic graph resembles
a feed-forward network albeit with a delay caused by the evaluation over a number of steps.

According to their definition, Echo Networks do not contain bias neurons 'outside' the network proper
in the way that conventional networks have bias nodes in addition to the nodes of the layer. This is 
foremost to keep all (genetic) information of the network in a single matrix. Since bias values
might be needed in small networks to avoid unnecessary processing steps/neurons they can be 
created implicitly by setting their columns in the connections matrix to $0$ (no other neuron has
them as a destination of a synapse) except for the value on the diagonal which needs to be set to $1$.   

Echo Networks were designed for discrete (time) signals for which usually the number of
sequential input data points is larger than the longest path through the network, both in training
and at inference. If this is not the case, Echo Networks can still be used, even for classification
tasks where there is no sequence at all, e.g., the classic Fisher's Iris dataset. The only requirement
is that the network is evaluated at least as often as the longest acyclic path through the network. 
Otherwise, some neurons and/or synapses might not contribute to the output result and the related 
computations would be wasted. Trivially, setting the number of evaluations to the number of neurons 
in the network would always suffice. Achieving the required number of evaluation steps can be
accomplished by either presenting the input repeatedly at each evaluation step or only once
at the beginning and then let the step-wise evaluation transport (and transform) them through the network. 
The latter, the phenomenon that input values will always be 'reflected' several times in any 
ordinary Echo Network before they vanish inspired its name.      

Since the post-activation values $\bm{a}$ are already used in the first evaluation step,
they need to be appropriately initialised. Setting them all to $0$ appears to be an obvious
choice but in our work we obtained better results by setting them to $1$.

\begin{figure}[htbp]
\centerline{\includegraphics[trim=250 120 250 120,clip,scale=0.5]{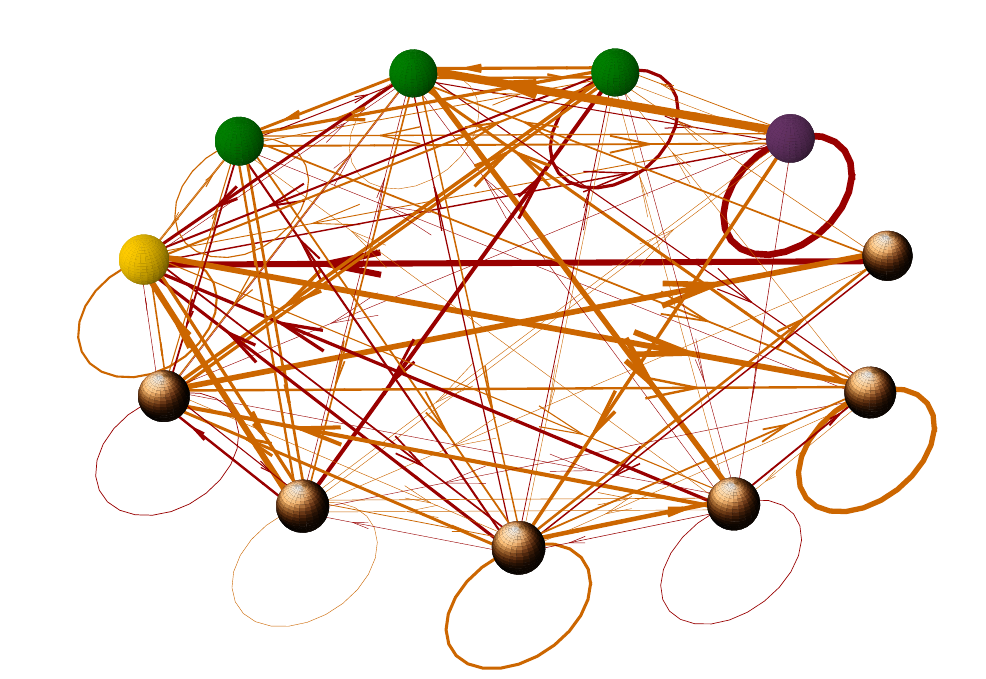}}
\caption{Sample of an Echo Network evolved through neuroevolution for the 
         classification of electrocardiography signals. The network reached an accuracy of $0.696$ on 
         the test set in the ECG classification task (see section~\ref{sec:datasets}) despite consisting 
         of only $11$ neurons and $121$ weights ($97$ non-zero).
         Designated input neurons: green; designated output neuron: yellow; bias neuron: 
         purple. A light hue indicates weights with negative sign, a dark hue 
         weights with positive sign. Line width is proportional to weight magnitude.
         }  
\label{fig:sample_Echo_Network}
\end{figure}

\section{Related work}
\label{sec:related_work}

There is a structural resemblance to reservoir computing 
\cite{verstraeten2007experimental} (e.g., liquid state machines \cite{maass2002real}, 
echo state networks \cite{jaeger2004harnessing}) in the way that information does 
not flow with a clear set path from the input layer via sequential latent layers 
to the output layer. But the similarities are superficial because Echo Networks
do not use random weights which are central to reservoir computing. 

The structural uniformity of a network is also known from Hopfield networks 
\cite{hopfield1982neural, krotov2016dense}. However, learning in 
Hopfield networks is energy-based, and they do not process
a temporal input sequence (although they can be adapted for the latter 
\cite{ramsauer2020hopfield}).  

The connection matrix has already been used in our previous neuroevolution research in 
\cite{kroos2017neuroevolution}, but only as an organising principle while layers were still 
determined post-hoc (after the creation of a new network through mutation and recombination) 
through computing the forward path length and assigning it as layer index. 
Accordingly, the computation of the activation states was based on sub-matrices consisting of 
the columns in the connection matrix of the neurons with the same layer index. It proceeded 
sequentially as in a conventional feedforward network or RNN.

\section{Experiments}
\label{sec:experiments}

\subsection{Dataset and task}
\label{sec:datasets}

To test Echo Networks in a realistic situation with real-world data we applied them
using neuroevolution for classification of electrocardiography (ECG) signals. We used
the PTB-XL dataset \cite{wagner2020ptb}, a large annotated clinical 12-lead ECG-waveform 
dataset, which consists of $21,837$ 10-seconds recordings from $18,885$ patients.
We selected a single channel (standard limb lead III, recording the potential 
difference between the left leg and the left arm) and set as task the binary 
classification of the recordings into 'normal' versus otherwise. 
Following the recommendation of the authors, subsets $1$ to $8$ were used for training,
subset 9 for validation during training and subset $10$ as test. However, 
$2174$ recordings were randomly selected and removed from the training set as an untouched 
set-aside dataset for future test use. Furthermore, we excluded recordings which were not 
human-annotated or for which the reported confidence of the cardiologist for the 
classification as 'normal' was below $50$\%. 

The dataset contains non-normal ECG 
recordings from a wide range of cardiovascular diseases and disorders organised according to 
diagnostic classes and meta-classes. We pooled all of them into a broad class with label 
'atypical' and balanced validation and test datasets to be able to look directly at 
accuracy (and true positives and negatives). Table~\ref{tab:datasets} shows the resulting 
numbers of recordings in each of the datasets.

\renewcommand{\arraystretch}{1.1}
\begin{table}[htbp]
\caption{Number of recordings in dataset split}
\begin{center}
\begin{tabular}{|l|r|r|r|}
\hline
Processing state & \textbf{\textbf\textit{Training}}& \textbf{\textit{Validation}}& \textbf{\textit{Test}} \\
\hline
Input & 15244 & 2198 & 2183 \\
Excluded, no human annotation & 5041 & 0 & 0 \\
Excluded, low confidence & 243 & 65 & 71 \\
Final unbalanced & 9960 & 2133 & 2112 \\
Normal & 4924 & 898 & 884 \\
Atypical & 5036 & 1235 & 1228 \\
Final balanced & - & 1796 & 1768 \\
\hline
\end{tabular}
\label{tab:datasets}
\end{center}
%\vspace*{-10pt}
\end{table}

The PTB-XL dataset offers each recording with the original $500$ Hz sample rate and a version
downsampled to $100$ Hz. The latter was used in this experiment and the data points of the
selected channel were arranged to contain 
all instances of $3$ consecutive samples that could be obtained without zero-padding, resulting
in a matrix of size $998$ x $3$. Each row of the matrix would become later the input of a 
single evaluation step of the network without further processing.

\subsection{Method}
\label{sec:method}
We used computational neuroevolution to evolve minimal networks for the ECG binary 
classification task, both with RNNs and Echo Networks. Starting point 
was the method used in our previous research \cite{kroos2017neuroevolution} (but without the 
spectral density transformations), which in turn is based on the NEAT algorithm 
\cite{stanley2002evolving} and further inspired by \cite{wang2013application}.
Details about the procedure can be found in the mentioned publications; however,
we substantially modified some aspects for better exploration and exploitation of
the search space. The method in \cite{kroos2017neuroevolution} was chosen
because it had been shown to be successful in acoustic event detection, a task similar
to detecting anomalies in ECG signals, which often occur only intermittently during 
a recording. However, in the ECG PTB XL dataset irregularities are not marked in the
signal but only the entire recording annotated, requiring further changes to the method.

The procedure can be summarised as follows:\\ 
A single population with $200$ individuals was let to evolve over $200$ generations. 
We used stochastic neuroevolution by selecting randomly in each generation
a subset of the recordings of the training data (set to $5$\%, amounting to 
$498$ recordings).

Each individual neural network was evaluated on the selected ECG recordings
by providing at \textit{each evaluation step} a triplet of consecutive raw data points to 
the three input neurons. Using the sigmoid function and rounding, the single output 
neuron of the network generated a binary value, with $0$ indicating a normal and $1$ an atypical signal. 
The resulting $998$ output values were averaged, and a threshold applied to determine whether the 
entire recording would be classified as normal. Using the inverse of the classification
error of all selected signals, a fitness value for each individual was computed.

Since the fitness of an individual in stochastic neuroevolution varies very strongly,  
the impact of this variation and oscillatory tendencies were smoothed by averaging the current 
fitness value with the one obtained from the previous generation. We used speciation 
\cite{stanley2002evolving}, which assigns networks sufficiently different from their peers to their 
own species and the fitness comparison for the selection process is only conducted within a species. 
Difference measures for speciation took into account topological and weight magnitude differences. 
We also used shared fitness \cite{stanley2002evolving} where the fitness of an individual is 
normalised by the number of members in its species.

The selection process was implemented by using stochastic universal sampling \cite{baker1987reducing}
for both mutation and recombination candidates based on their fitness values. The elimination proportion 
was set $66$\%. In addition, we used elitism \cite{dejong1975analysis}: The top $6$ fittest individuals 
were included in the next generation without any changes. For synapse changes, 
both drawing a new value from a Gaussian distribution and Gaussian perturbation of the current 
weight were used as mutation operators with an equal probability of $0.5$. Synapses and neurons 
could be added to or removed from the network as consequence of a mutation. As recombination 
operators, both crossover and weight averaging were applied with an equal probability of $0.5$.

The evolution process started with the first generation of minimal networks containing only the 
three input neurons, the output neuron and a bias neuron plus the forward and recurrent synapses
between them in the RNN and, equivalently, the bidirectional connections in the Echo Network. 
The activation function was chosen to be ReLU for RNNs as well as Echo Networks
and was not subjected to mutations. The input functions in the Echo Networks were set
for two neurons to the identity function and one neuron to a sign reversal.   
At the end of each generation the best network (highest fitness value) was evaluated on the entire
validation set to track the generalizability of the evolved solution across generations. The network
was stored if it achieved a lower error on the validation set than the best network of the previous 
generations. After completion of the final generation, this network was applied on the test set to 
arrive at the final performance value.
%The entire evolution process was run on a single core of a XYZ CPU of an HP workstation. It took
%approximately X min in wall clock time. 
Each run was repeated ten times to get a coarse estimate of the involved random variability. 

In general, it can be assumed with neuroevolution that further improvements could be achieved by 
scaling up the evolution process (e.g., more populations, more individuals). To confirm this for
Echo Networks we employed $8$ populations of $120$ individuals using 'islands' (exchange of individuals 
every $n$ generations, with $n$ set to $4$) in a second experiment.

\subsection{Results}
\label{sec:results}
Since we employed a binary classification and had an equal number of observations
in both classes in the validation and test sets, accuracy is a sufficient measure.

\renewcommand{\arraystretch}{1.2}
\begin{table}[htbp]
\caption{Performance results from 10 evolution runs.}
\begin{center}
\begin{tabular}{|l|c|c|c|c|}
\hline
\textbf{Network type}&\multicolumn{4}{|c|}{\textbf{Accuracy}} \\
\cline{2-5} 
 & \textbf{\textit{Mean}}& \textbf{\textit{Std}}& \textbf{\textit{Min}} & \textbf{\textit{Max}} \\
\hline
RNN & 0.671 & 0.007 & 0.662 & 0.684 \\
Echo Network & 0.687 & 0.005 & 0.679 & 0.696 \\
\hline
\end{tabular}
\label{tab:results}
\end{center}
\end{table}

Table~\ref{tab:results} shows the results. Evolved Echo Networks perform slightly
better than evolved RNNs. The Echo Network with the best performance is shown in 
Fig.~\ref{fig:sample_Echo_Network}. 

\begin{figure}[htbp]
\centerline{\includegraphics[trim=280 120 270 120,clip,scale=0.5]{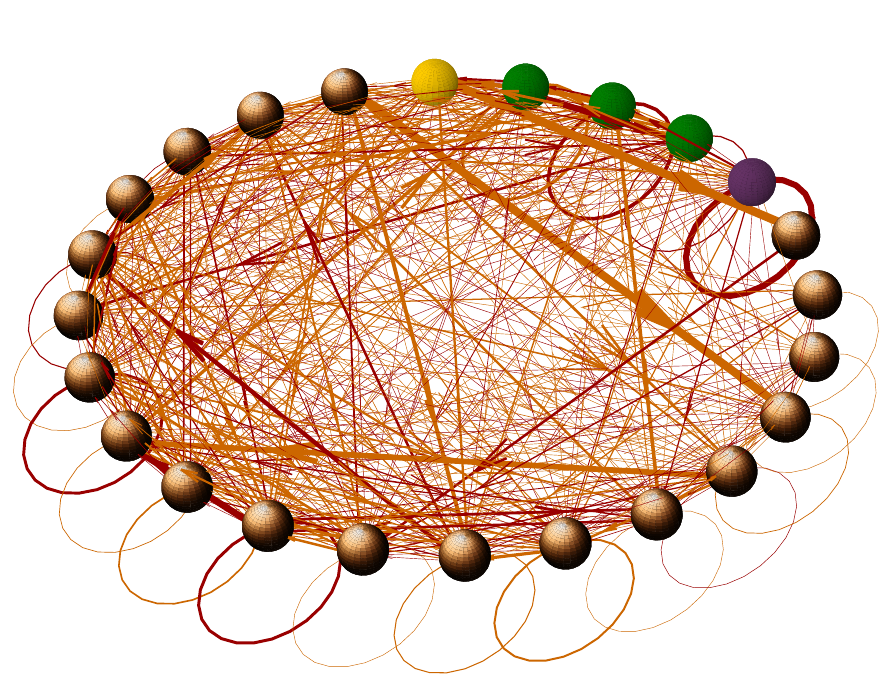}}
\caption{The network reached an accuracy of $0.717$ on 
         the test set in the ECG classification task (see section~\ref{sec:datasets}) 
         It has $24$ neurons and $576$ weights ($447$ non-zero).
         Legend is the same as for Fig.~\ref{fig:sample_Echo_Network}.
         }  
\label{fig:Echo_Network_8_pop_run}
\end{figure}

The second experiment with $8$ populations and only Echo Networks
resulted in a mean accuracy of $0.701$ with a 
standard deviation of $0.009$ and a maximum of $0.717$. The network with the best performance is 
shown in Fig.~\ref{fig:Echo_Network_8_pop_run}. Note that the cut-off at $200$ generations in both
experiments is early and was set to keep the networks small enough for visualisation in the
current paper. 

\begin{figure}[htbp]
\centerline{\includegraphics[trim=0 0 0 0,clip,scale=0.45]{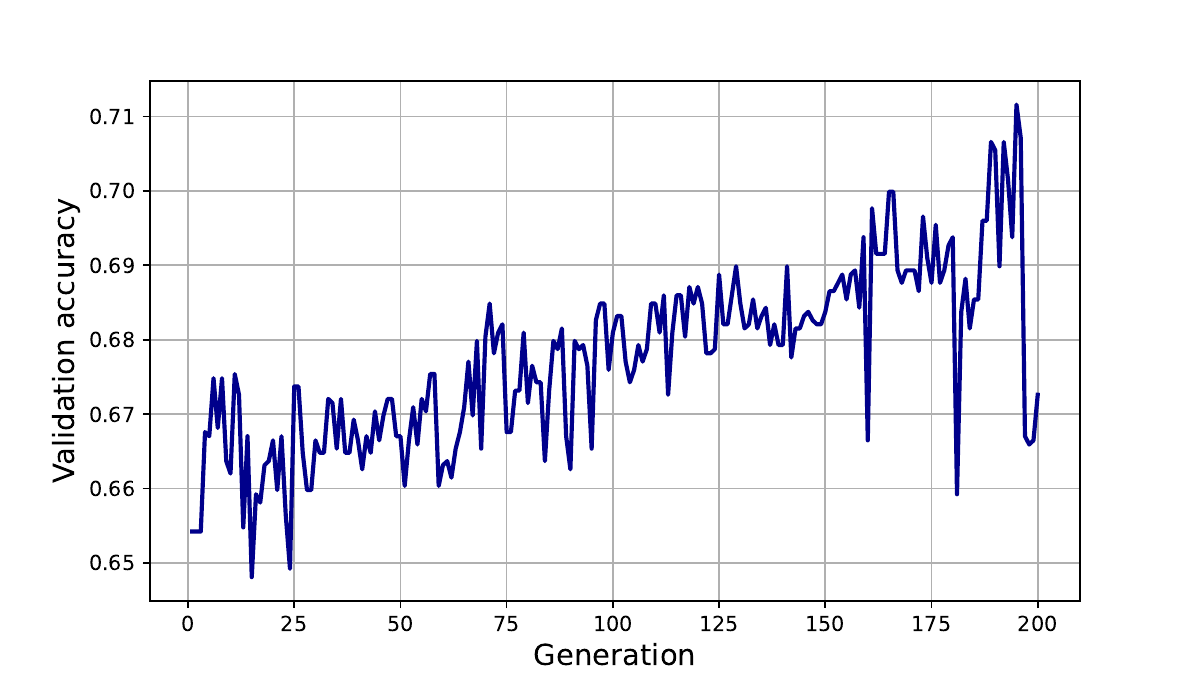}}
\caption{Validation accuracy across generations resulting in the network 
         shown in Fig.~\ref{fig:Echo_Network_8_pop_run} at generation $195$.
         }  
\label{fig:Echo_Network_8_pop_run_val_results}
\end{figure}

Fig.~\ref{fig:Echo_Network_8_pop_run_val_results} tracks the development of the validation
results over the course of the $200$ generations. At each 
generation, the training data consists of only $5$\% of the entire training dataset and only
the network achieving the highest fitness on the subset is evaluated on the validation set.
This explains the high variability which is to be expected when using stochastic neuroevolution
and works as a regulation mechanism.

\section{Discussion}
\label{sec:dicussion}
The experiment results hint at the potential power
of Echo Networks but clearly more evaluations and comparisons
are needed. We see the primary benefits of Echo Networks in the
simplification of the evolution process and how they enable a more 
systematic approach to mutation and
recombination. In contrast to conventional networks, they do not require
historical markers to identify neurons for adequate recombination as in NEAT
or the computationally intense full topological analysis which would be the
alternative. Since adding a neuron as mutation outcome takes the form of
adding a row and a column to the connection matrix, the size of the matrix
becomes the decisive topological difference criterion followed by the distribution 
of entries with non-zero weights. How these emerged in the evolution process
is of no significance: Networks with the same connection matrices are indeed identical, 
and it stands to reason that networks with \textit{similar} connection matrices show 
\textit{similar} inference performance and achieve it in similar ways, though this 
still needs to be shown.   

The representation of the network as square matrix enables the use of matrix computations 
and factorisations in mutation and recombination. This might help
to solve a critical problem of direct genetic encoding in neuroevolution: In general,
the combination of two high performing but topologically different parent networks 
does not lead to child networks with similar good or even better performance. 
Most of the time it is the opposite even if the child network is given time 
to adjust the weights through speciation. This shortcoming of direct encoding 
impacts negatively one of its biggest advantages, namely, that small changes in the 
genetic code lead to small changes in the network structure. However, these small
changes might still cause large differences in the performance of the network. A more principled,
theory-guided approach and mutation and recombination operators based 
on the connection matrix could alleviate this problem.  

The connection matrix grows quadratically with the number of neurons but 
this is not different from the weight matrix of fully connected MLP layers with an
equal number of source and destination neurons. Furthermore, many of the methodological
decisions are currently guided by heuristics accumulated over previous research in the 
entire field. None of them are specifically tested for Echo Networks. Meta-parameters settings, 
if not taken from the literature, are often determined by trial and error for
a given task. 

\section{Conclusion and outlook}
\label{sec:conclusion_and_outlook}
We introduced a new type of network named Echo Network, which
promises the generation of powerful minimal networks via neuroevolution.
Through its definition based on the connection matrix and its flexible
assignment of input and output neurons it enables 
a more systematic approach to mutation and recombination and a theory-guided analyses
of the solution space for a given task. Preliminary evidence
from the classification of ECG signals shows that Echo Networks are able to
compete with conventional feed-forward networks and RNNs. 

Future work will include an extension of the empirical evaluation of Echo 
Networks, in particular, by applying them with neuroevolution to a broad range 
of tasks and data.  
Furthermore, Echo Networks are not confined to the use with neuroevolution, but
can also be trained like conventional networks - via gradient descent and 
backpropagation \cite{rumelhart1985learning}, by using the forward 
gradient \cite{baydin2022gradients} or via the forward-forward
algorithm \cite{hinton2022forward}. How well they will fair is currently unclear. 
%For instance, for the training with backpropagation they might need additional 
%mechanisms equivalent to skip connections or gating outside the connection matrix. 

Finally, Echo Networks 
can be scaled up to larger composite networks. One approach to accomplish this would
be to construct in a strict recursive fashion: Instead of having neurons as their basic elements, 
higher-level Echo Networks would contain Echo Networks themselves. Another approach would
consist of irregular nettings of several Echo Networks of different sizes without a hierarchical
structure, probably combined with a dynamic routing mechanism using the key property of 
Echo Networks that input and output neurons are not structurally fixed.

\bibliographystyle{IEEEtran}
\bibliography{IEEEabrv,echo_nets}

@article{kroos2017neuroevolution,
  title={Neuroevolution for sound event detection in real life audio: A pilot study},
  author={Kroos, Christian and Plumbley, Mark},
  journal={Detection and Classification of Acoustic Scenes and Events (DCASE 2017) Proceedings 2017},
  year={2017},
  publisher={Tampere University of Technology}
}

@article{wagner2020ptb,
  title={{PTB-XL}, a large publicly available electrocardiography dataset},
  author={Wagner, Patrick and Strodthoff, Nils and Bousseljot, Ralf-Dieter and Kreiseler, Dieter and Lunze, Fatima I and Samek, Wojciech and Schaeffter, Tobias},
  journal={Scientific Data},
  volume={7},
  number={1},
  pages={1--15},
  year={2020},
  publisher={Nature Publishing Group}
}

@article{wang2013application,
  title={The application of improved {N}euro{E}volution of {A}ugmenting {T}opologies neural network in {M}arcellus {S}hale lithofacies prediction},
  author={Wang, Guochang and Cheng, Guojian and Carr, Timothy R},
  journal={Computers \& Geosciences},
  volume={54},
  pages={50--65},
  year={2013},
  publisher={Elsevier}
}

@article{stanley2002evolving,
  title={Evolving neural networks through augmenting topologies},
  author={Stanley, Kenneth O and Miikkulainen, Risto},
  journal={Evolutionary Computation},
  volume={10},
  number={2},
  pages={99--127},
  year={2002},
  publisher={MIT Press}
}

@techreport{rumelhart1985learning,
  title={Learning internal representations by error propagation},
  author={Rumelhart, David E and Hinton, Geoffrey E and Williams, Ronald J},
  institution={Institute for Cognitive Science, University of California, San Diego},
  year={1985}
}

@article{baydin2022gradients,
  title={Gradients without backpropagation},
  author={Baydin, At{\i}l{\i}m G{\"u}ne{\c{s}} and Pearlmutter, Barak A and Syme, Don and Wood, Frank and Torr, Philip},
  journal={arXiv preprint arXiv:2202.08587},
  year={2022}
}

@article{hinton2022forward,
  title={The forward-forward algorithm: {S}ome preliminary investigations},
  author={Hinton, Geoffrey},
  journal={arXiv preprint arXiv:2212.13345},
  volume={2},
  number={3},
  pages={5},
  year={2022}
}

@article{maass2002real,
  title={Real-time computing without stable states: {A} new framework for neural computation based on perturbations},
  author={Maass, Wolfgang and Natschl{\"a}ger, Thomas and Markram, Henry},
  journal={Neural omputation},
  volume={14},
  number={11},
  pages={2531--2560},
  year={2002},
  publisher={MIT Press}
}

@article{jaeger2004harnessing,
  title={Harnessing nonlinearity: {P}redicting chaotic systems and saving energy in wireless communication},
  author={Jaeger, Herbert and Haas, Harald},
  journal={Science},
  volume={304},
  number={5667},
  pages={78--80},
  year={2004},
  publisher={American Association for the Advancement of Science}
}

@article{verstraeten2007experimental,
  title={An experimental unification of reservoir computing methods},
  author={Verstraeten, David and Schrauwen, Benjamin and d’Haene, Michiel and Stroobandt, Dirk},
  journal={Neural Networks},
  volume={20},
  number={3},
  pages={391--403},
  year={2007},
  publisher={Elsevier}
}

@article{stanley2007compositional,
  title={Compositional pattern producing networks: A novel abstraction of development},
  author={Stanley, Kenneth O},
  journal={Genetic Programming and Evolvable Machines},
  volume={8},
  number={2},
  pages={131--162},
  year={2007},
  publisher={Springer}
}

@article{whitelam2021correspondence,
  title={Correspondence between neuroevolution and gradient descent},
  author={Whitelam, Stephen and Selin, Viktor and Park, Sang-Won and Tamblyn, Isaac},
  journal={Nature Communications},
  volume={12},
  number={1},
  pages={6317},
  year={2021},
  publisher={Nature Publishing Group UK London}
}

@inproceedings{baker1987reducing,
  title={Reducing bias and inefficiency in the selection algorithm},
  author={Baker, James E and others},
  booktitle={Proceedings of the Second International Conference on Genetic Algorithms},
  volume={206},
  pages={14--21},
  year={1987}
}

@book{dejong1975analysis,
  title={An analysis of the behavior of a class of genetic adaptive systems},
  author={De Jong, Kenneth Alan},
  year={1975},
  publisher={University of Michigan}
}

@book{dejong2006evolutionary,
  title={Evolutionary Computation: {A} Unified Approach},
  author={De Jong, Kenneth Alan},
  year={2006},
  publisher={The MIT Press}
}

@article{stanley2019designing,
  title={Designing neural networks through neuroevolution},
  author={Stanley, Kenneth O and Clune, Jeff and Lehman, Joel and Miikkulainen, Risto},
  journal={Nature Machine Intelligence},
  volume={1},
  number={1},
  pages={24--35},
  year={2019},
  publisher={Nature Publishing Group UK London}
}

@InProceedings{pmlr-v38-choromanska15,
  title = 	 {{The Loss Surfaces of Multilayer Networks}},
  author = 	 {Choromanska, Anna and Henaff, MIkael and Mathieu, Michael and Ben Arous, Gerard and LeCun, Yann},
  booktitle = {Proceedings of the Eighteenth International Conference on Artificial Intelligence and Statistics},
  pages = 	 {192--204},
  year = 	 {2015},
  editor = 	 {Lebanon, Guy and Vishwanathan, S. V. N.},
  volume = 	 {38},
  series = 	 {Proceedings of Machine Learning Research},
  address = 	 {San Diego, California, USA},
  publisher =    {PMLR},
}

@article{hopfield1982neural,
  title={Neural networks and physical systems with emergent collective computational abilities.},
  author={Hopfield, John J},
  journal={Proceedings of the National Academy of Sciences},
  volume={79},
  number={8},
  pages={2554--2558},
  year={1982}
}

@article{krotov2016dense,
  title={Dense associative memory for pattern recognition},
  author={Krotov, Dmitry and Hopfield, John J},
  journal={Advances in Neural Information Processing Systems},
  volume={29},
  year={2016}
}

@article{ramsauer2020hopfield,
  title={Hopfield networks is all you need},
  author={Ramsauer, Hubert and Sch{\"a}fl, Bernhard and Lehner, Johannes and Seidl, Philipp and Widrich, Michael and Adler, Thomas and Gruber, Lukas and Holzleitner, Markus and Pavlovi{\'c}, Milena and Sandve, Geir Kjetil and others},
  journal={arXiv preprint arXiv:2008.02217},
  year={2020}
}

\end{document}